\documentclass[conference]{IEEEtran}
\IEEEoverridecommandlockouts
\usepackage{amsmath,amssymb,amsfonts}
\usepackage{algorithmic}
\usepackage{graphicx}
\usepackage{textcomp}
\usepackage{xcolor}

\usepackage{multirow}
\usepackage{hyperref}
\usepackage{verbatim}
\usepackage{placeins}

\def\BibTeX{{\rm B\kern-.05em{\sc i\kern-.025em b}\kern-.08em
    T\kern-.1667em\lower.7ex\hbox{E}\kern-.125emX}}
    
\begin{document}

\title{CharNeRF: 3D Character Generation from Concept Art}

\author{

\IEEEauthorblockN{Eddy Chu}
\IEEEauthorblockA{\textit{National University of Singapore} \\
Singapore \\
e0418218@u.nus.edu}

\and
\IEEEauthorblockN{Yiyang Chen}
\IEEEauthorblockA{\textit{National University of Singapore} \\
Singapore \\
e0563851@u.nus.edu}

\and
\IEEEauthorblockN{Chedy Raissi}
\IEEEauthorblockA{\textit{Riot Games} \\
Singapore \\
chedy.raissi@inria.fr}

\and
\IEEEauthorblockN{Anand Bhojan}
\IEEEauthorblockA{\textit{National University of Singapore} \\
Singapore \\
dcsab@nus.edu.sg}

}

\maketitle

\begin{abstract}
3D modeling holds significant importance in the realms of AR/VR and gaming, allowing for both artistic creativity and practical applications. However, the process is often time-consuming and demands a high level of skill. In this paper, we present a novel approach to create volumetric representations of 3D characters from consistent turnaround concept art, which serves as the standard input in the 3D modeling industry. While Neural Radiance Field (NeRF) has been a game-changer in image-based 3D reconstruction, to the best of our knowledge, there is no known research that optimizes the pipeline for concept art. To harness the potential of concept art, with its defined body poses and specific view angles, we propose encoding it as priors for our model. We train the network to make use of these priors for various 3D points through a learnable view-direction-attended multi-head self-attention layer. Additionally, we demonstrate that a combination of ray sampling and surface sampling enhances the inference capabilities of our network. Our model is able to generate high-quality 360-degree views of characters. Subsequently, we provide a simple guideline to better leverage our model to extract the 3D mesh. It is important to note that our model's inferencing capabilities are influenced by the training data's characteristics, primarily focusing on characters with a single head, two arms, and two legs. Nevertheless, our methodology remains versatile and adaptable to concept art from diverse subject matters, without imposing any specific assumptions on the data.
\end{abstract}


\begin{figure*}[h]
    \centering
    \includegraphics[width=1\textwidth]{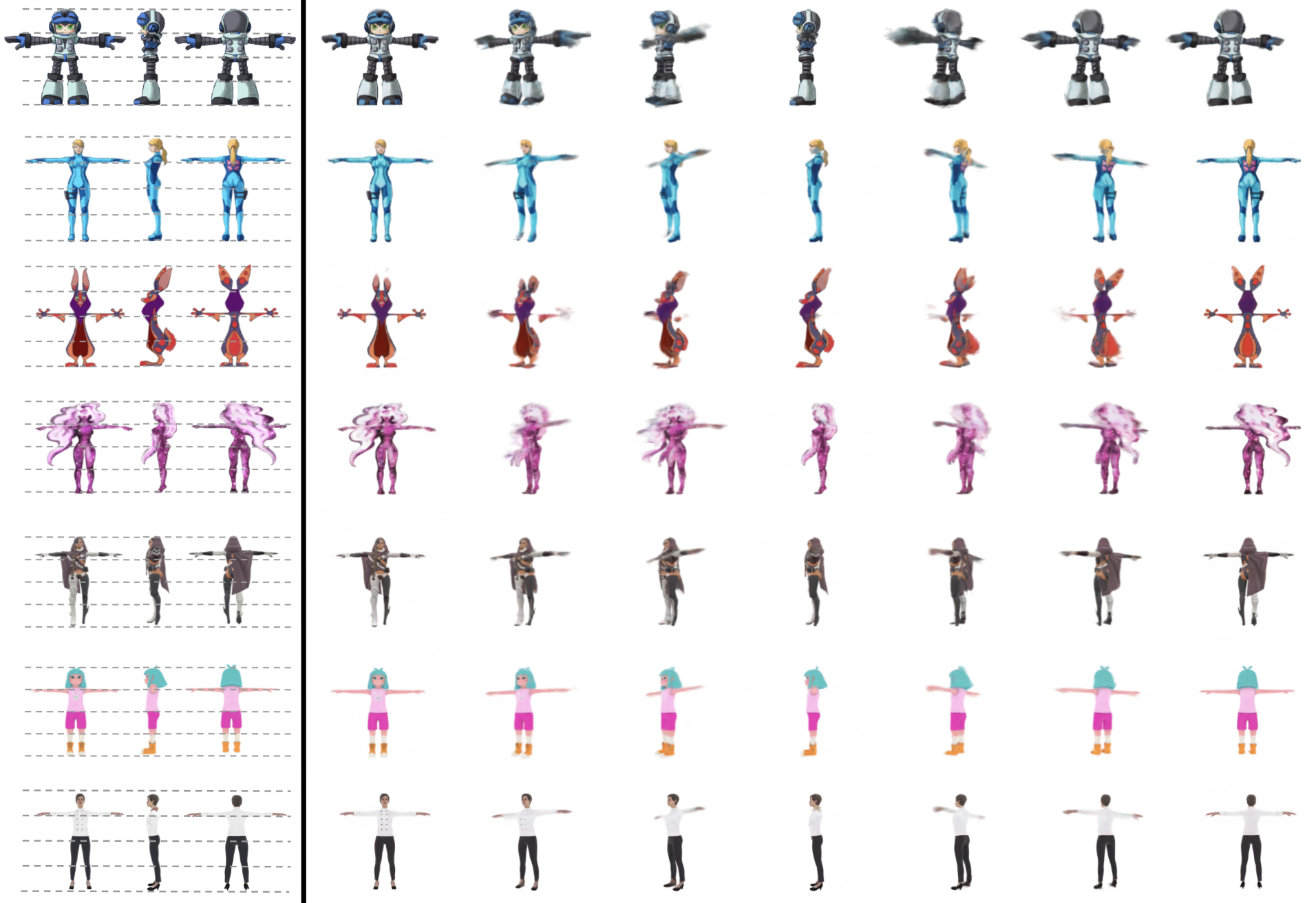}
    \caption{ \textnormal{CharNeRF constructs neural radiance field for characters from concept art (front, side and back images in the left columns). CharNeRF is able to infer novel views of characters with complicated shapes. (in the right columns) First four rows are results for real concept arts created by 2D artists, while the last 3 rows use synthetic inputs created from existing 3D mesh.} }
    \label{fig:intro:teaser}
\end{figure*}

\section{Introduction}
Three-dimensional (3D) modeling plays a pivotal role in AR/VR and gaming, offering versatile assets for educational simulations, artistic installations, engaging games and metaverse. The conventional approach to crafting 3D models involves utilizing software like Blender or Autodesk Maya, which can be incredibly time-consuming and demands a high level of expertise. In fact, statistics reveal that it often requires hundreds of hours for a skilled artist to create a single high-quality 3D model for a visually appealing asset~\footnote{\href{https://wallawallastudio.com/article/how-long-does-it-take-to-make-a-3d-character-model/}{https://wallawallastudio.com/article/how-long-does-it-take-to-make-a-3d-character-model/}}. Recent advancements in procedural content generation through deep learning techniques have opened up opportunities for the automated creation of various game assets, including game maps~\cite{summerville_procedural_2018}. This academic undertaking serves a dual purpose: it contributes to the scholarly discourse and seeks to alleviate the arduous process of manually modeling 3D characters. The core objective of this project is to explore and develop a procedural content generation framework that can autonomously generate 3D representations for characters, aiming to significantly reduce the labor and expertise required in this intricate field.

This paper addresses the specific challenge of generating 3D characters from 2D concept art. Concept art occupies a critical role within the development process, residing between the initial project ideas and the eventual final product. It can broadly be categorized into three distinct types: character concept art, world concept art, and equipment concept art~\footnote{\href{https://www.domestika.org/en/blog/5703-what-is-a-turnaround-in-character-design-and-how-to-draw-one}{https://www.domestika.org/en/blog/5703-what-is-a-turnaround-in-character-design-and-how-to-draw-one}}. In this paper, we narrow our focus to character concept art, which ranges from simple uncolored sketches to dynamic action scenes and precise, consistent turnaround drawings. A 3D modeling-ready concept art demands meticulous consistency, often consisting of three to five distinct turnaround views. Specifically, the concept art outlined in this paper is defined to be the three-view turnaround drawings, including the front, side, and back perspectives, and these drawings typically feature characters posed in the standard T or A configuration. While it is important to acknowledge that the generation of turnaround concept art can pose significant challenges, it's worth emphasizing that our project's primary focus lies elsewhere. Notably, there are existing solutions, like "charturner"~\footnote{\href{https://civitai.com/models/3036/charturner-character-turnaround-helper-for-15-and-21}{https://civitai.com/models/3036/charturner-character-turnaround-helper-for-15-and-21}}, that have been developed to address this particular task.

The problem we aim to address focuses on 3D reconstruction and aligns closely with industrial practices. VR/Game designers commonly start by crafting concept art for characters, often utilizing the default T or A pose. Subsequently, they hand over these concept arts to modeling specialists to create the 3D models for the characters. Our research delves into streamlining this process, with a focus on character concept art and the generation of 3D models from these initial artistic representations.

3D model reconstruction represents a well-explored problem, but the techniques commonly used in this field often cannot be readily applied to our specific domain of virtual characters. For instance, the ability to accurately infer 3D models already exists for domains such as faces and clothed humans, thanks to the use of high-precision capture systems~\cite{highQualitySteamable, theRelightables}. Regrettably, these sophisticated capture systems are not available for use in the virtual world. In recent research on 3D model reconstruction from 2D images of clothed humans, there have been significant advancements that enable the prediction of 3D human figures with reasonable surface details, often requiring only a limited number of real-world images~\cite{saito2020pifuhd, icon, AnimationReadyClothedHuman, siclope}. However, it's important to note that these approaches often rely on "human shape assumptions" within their pipelines, which makes their generalization or transferability to other source domains, such as animals or virtual characters, a challenging task. Common approaches within this context frequently revolve around the utilization of well-established or learned human models, including the SMPL model~\cite{smpl}, GHUM model~\cite{ghum}, and SCAPE model~\cite{scape}. Unfortunately, these models are typically not available or suitable for use with virtual characters, underlining the unique complexities of our target domain.

In this paper, we confront the challenge using the neural radiance field (NeRF)\cite{nerf}. NeRF offers a distinct advantage in terms of rendering resolution, surpassing the limitations of methods reliant on volumetric voxel representations like Deephuman and BodyNet\cite{Deephuman, bodyNet}, thanks to its memory-efficient implicit nature. Notably, volumetric representation of 3D models provides a flexible rendering resolution, a valuable asset for VR or AR, where various levels of detail (LOD) in 3D mesh are required for efficient rendering. To bolster this approach, we incorporate an image encoder, extracting pixel-aligned information from the concept art and combining these meaningful representations through a view direction-attended multi-head self-attention layer. This approach essentially emulates the way a human artist creates a 3D model, inferring depth from nearby drawings. Expanding beyond the neural radiance field pipeline, we enhance rendering quality by employing a fusion of sampling methods that simultaneously learn to render new views and reconstruct shapes. To gauge the effectiveness of our approach, we generate fresh views for 3D characters and validate them with pictures taken of the characters. The front, side, and back images serve as concept art inputs, and the availability of ground truth 3D models allows for an assessment of the quality of novel view reconstruction. Given the distinctiveness of our work within the neural radiance field in this innovative domain, we also conduct an ablation study to further validate the efficacy of our approach, and assess the effectiveness of our method using real concept art drawings. Please refer to Figure \ref{fig:intro:teaser} for a glimpse of the novel views of virtual characters synthesized by our model with concept art as input.

In summary, our \textbf{contributions} are as follows: (1) CharNeRF: We introduce CharNeRF, a model capable of inferring detailed novel views of 3D character's concept art - the first to target 3D game/VR/AR characters with clear definition of concept art. (2) Feature Vector Combination: We present a novel method for combining feature vectors by learning feature and view similarity, enhancing the quality of our results. (3) Mix of Sampling Methods: Our approach includes training the network using a mix of sampling methods, capitalizing on data similarity within the dataset, thereby improving rendering quality. (4) Mesh Reconstruction Guidelines: We offer practical guidelines for mesh reconstruction that optimize the capabilities of our network.

\section{Related Work}
\textbf{NeRF based Reconstruction} 
The neural radiance field is a volumetric 3D representation that represents a scene as a function taking in both 3D location and 2D viewing direction as inputs $(\theta, \phi)$ (solid angle)\cite{nerf} and returning the volume density and emitted colour. NeRF models show impressive results for scene construction with high surface details shown. However, the original NeRF model and subsequent works require dense inputs~\cite{nerf, blockNerf} and have a long optimisation time. This makes NeRF models hard to scale for our purpose - lightweight 3D model reconstruction from few-shot 2D images. Although there are methods that speed up NeRF training processes by improving internal data representation for NeRF~\cite{instant-ngp, fridovich-keil_plenoxels_2022}, such methods still require huge amounts of input images. Another line of research~\cite{keypointNerf, portraitNerf, pixelnerf, johari2022geonerf, humannerf, su_-nerf_2021} in NeRF is to introduce priors and turn the model into a supervised one. Such methods typically include a convolutional image encoder and make the NeRF component process information from the encoded feature maps instead of directly from the input images. Some also include additional information such as ground truth or predicted depth values~\cite{johari2022geonerf, keypointNerf} or impose constraints and regularisation terms on the NeRF model~\cite{kim2022infonerf, niemeyer2022regnerf}. 

Given the specific context of our problem, we opt to adopt the pixelNeRF framework~\cite{pixelnerf} as the foundational model for our work. Notably, pixelNeRF exhibits effectiveness without the need for additional priors beyond the input images, making it a promising choice for few-shot reconstruction tasks.

The architectures of few-shot NeRF models are explicitly designed to facilitate scene reconstruction from a minimal number of images, often a single image or a handful of them. These models often incorporate multi-view reasoning techniques, which involve amalgamating intermediate NeRF representations derived from each input view to arrive at a final prediction encompassing multiple views. Common approaches include straightforward average pooling of intermediate features for the same pixel across different views, as seen in the pixelNeRF model~\cite{pixelnerf}. Alternatively, some models harness self-attention mechanisms~\cite{self-attention} to enrich the information exchange among features from distinct views.

Typically NeRF models are used to construct a general scene with one or multiple objects. However, there are NeRF frameworks that reconstruct a real-world human from images,~\cite{su_-nerf_2021, humannerf} and they also incorporate real human-related priors. For example, A-NeRF~\cite{su_-nerf_2021} uses an off-the-shelf SMPL~\cite{smpl} based sub-network to extract human pose and keypoint information from the images before passing them to NeRF. Given the inherently diverse shapes, colors, and artistic elements characterizing characters, we contend that it may be more prudent to treat each character as a distinct scene and leverage generic NeRF-based methodologies, as opposed to leaning heavily on domain-specific, learning-based models.

\textbf{Clothed Human Reconstruction} A relevant topic in 3D computer vision that has seen many breakthroughs recently is clothed human reconstruction which is to reconstruct the 3D representation of a real-world human from one or few photographs. Recent methods in this field can be broadly categorised into two types, depending on whether they rely on real-world human priors.

The first category comprises parametric models that harness existing parametric models as prior knowledge to bridge the gap between 2D input images and the ultimate 3D model. The model construction problem is thereby morphed into a parameter learning problem~\cite{icon, endToEndRecovery, zheng_pamir_2022}. The parametric model for humans has been studied intensively. Models such as SMPL\cite{smpl}, GHUM\cite{ghum}, and SCAPE\cite{scape} already describe naked humans in many sorts of poses well. From papers proposing parametric model based approaches, we observe accurate pose estimation from the final rendered 3D model but these approaches fail to reconstruct the surface details faithfully. This is because parametric models have limiting expressive power on the resulting model. Parametric model based approaches greatly simplify the 3D reconstruction problem because parameter learning is a well-studied area, but they fail to reconstruct the surfaces of clothed humans faithfully due to the complexity of the shapes of clothes and the limited expressiveness of the parametric model. 

The second category of approaches does not rely on any human body shape and pose priors and considers the task as a generic 3D object reconstruction from 2D images, under the implicit assumption that all objects are of the same domain, namely, human bodies. Models such as BodyNet~\cite{bodyNet}, PIFu~\cite{pifu}, and PIFuHD~\cite{saito2020pifuhd} fall under this category, and they reconstruct the model in different representations including voxel or implicit functions. Methods in this category can represent free-form objects but they rely heavily on huge amounts of training data for the models to learn. Unfortunately, the estimation for virtual game characters is a much tougher problem due to the scarcity of training data - unlike existing large-scale datasets for clothed humans such as Deepfashion~\cite{liu2016deepfashion} and RenderPeople~\cite{pifu}, to the best of our knowledge, there have not been benchmark datasets on 3D characters. Furthermore, the higher degree of freedom regarding its surface shape and drawing styles of game characters make such models even harder to generalise. Due to such issues, clothed human reconstruction models are not chosen over NeRF based approaches for our problem.

\section{Background}
Our approach is greatly based on neural radiance field ~\cite{nerf}. In this section, we briefly explain the methodology for and notations of NeRF and then we will discuss our method in Method section. (Section \ref{method})

Neural Radiance Field (NeRF) \cite{nerf} is a neural 3D representation. It represents a scene as a differentiable function \emph{f} that takes in positional encoded 3D location $x \in \mathbb{R}^3$ and a unit viewing direction $ d \in \mathbb{R}^3$ to predict the emitted color c = $(\emph{r}, \emph{g}, \emph{b})$ and volume density $\sigma$ of a given point viewed from a specific direction. Positional encoding is a mapping function that maps input to a higher dimensional representation with different frequencies. This technique is found to help NeRF capture color and geometry details of different frequencies better. Given the number of frequency $\nu$, a given 3D location $x \in \mathbb{R}^3$  is mapped to $\gamma(x) \in \mathbb{R}^{3*\nu*2}$ using positional encoding:
\begin{equation}
\label{positional_encode}
\gamma(x) = (\sin(2^0\pi x), \cos(2^0\pi x), ..., \sin(2^{\nu-1}\pi x), \cos(2^{\nu-1} \pi x))
\end{equation}

\noindent
, and the final expression for NeRF would be:
\begin{equation}
f(\gamma(x), d) = (c, \sigma)
\end{equation}

NeRF models are typically represented by Multi-Layer Perceptrons (MLPs) because of their robust ability to approximate high-dimensional functions~\cite{hornik_multilayer_1989}. Training a Neural Radiance Field for a scene necessitates images with known poses, including extrinsic and intrinsic matrices. During the training phase, the camera positions for each image can be deduced from the extrinsic matrix. Rays are emitted from each camera and directed toward every pixel in the image. Along each ray, points are initially sampled at even intervals, bounded by near and far values. These sample points are then input to the NeRF model to estimate the emitted color (c) and volume density ($\sigma$). This initial sampling process is often referred to as "coarse sampling." Utilizing the volume density values obtained in coarse sampling, a weighting value can be computed for each sample point, guiding further sampling based on the distribution of these weights. You can refer to the formula for this weightage in Equation \ref{weightage}. This subsequent sampling process is commonly known as "fine sampling."

Along each ray $r$, the model calculates the final rendered color $\hat{C}(r)$ by combining the estimated colors (c) and volume density ($\sigma$) using the following formula~\cite{nerf}:
\begin{equation} 
\label{colorcompositing}
\hat{C}(r) = \int_{t_n}^{t_f} T(t) \sigma(r(t)) c(x=r(t), d)\,dt 
\end{equation}

\noindent
, where $T(t) = \exp(-\int_{t_n}^{t} \sigma(r(s))\,ds$ and $t_n$ and $t_f$ represents the parameterized near and far value. \\

In practice, a discrete form of this calculation is provided below. In this form, \emph{n} represents the total number of sample points, $\alpha_i$ and $c_i$ stand for the estimated alpha value and predicted RGB value for each individual point i, and $T_i$ represents the transmittance up to the point i. It's important to note that these sample points are organized in ascending order based on their distance from the camera.
\begin{equation} 
\label{color-composite}
\hat{C}(r) \approx \Sigma_{i=1}^n T_i\alpha_i c_i
\end{equation}

\begin{equation} 
\label{weightage}
w_i = T_i\alpha_i 
\end{equation}
\label{integral-estimate}

\noindent
, where $\alpha_i$ can be derived from volume density $\sigma_i$ by $\alpha_i = \exp(-\sigma_i \delta_i)$ and the distance between nearby sample points $\delta_i = p_{i+1} - p_i$. As for the transmittance value $T_i$, it is computed through the alpha compositing algorithm given by $T_i = \Pi_{j=1}^{i-1} (1 - \alpha_i)$. 

Due to the fact that the whole rendering process is differentiable and the ray sampling process is equivalent to sampling pixels from images, the neural radiance field can be simply optimized by minimizing a simple L2 loss between the ground truth pixel value and the composited rgb value.
\begin{equation} 
    L_\text{reconstruction} = \frac{1}{|\mathcal{R}|} \Sigma_{r \in \mathcal{R}} \|\hat{C}(r) - C(r)\|^2_2
\end{equation}

\smallskip

\section{Methods}
\label{method}

\begin{figure}[h]
    \centering
    \includegraphics[width=9cm]{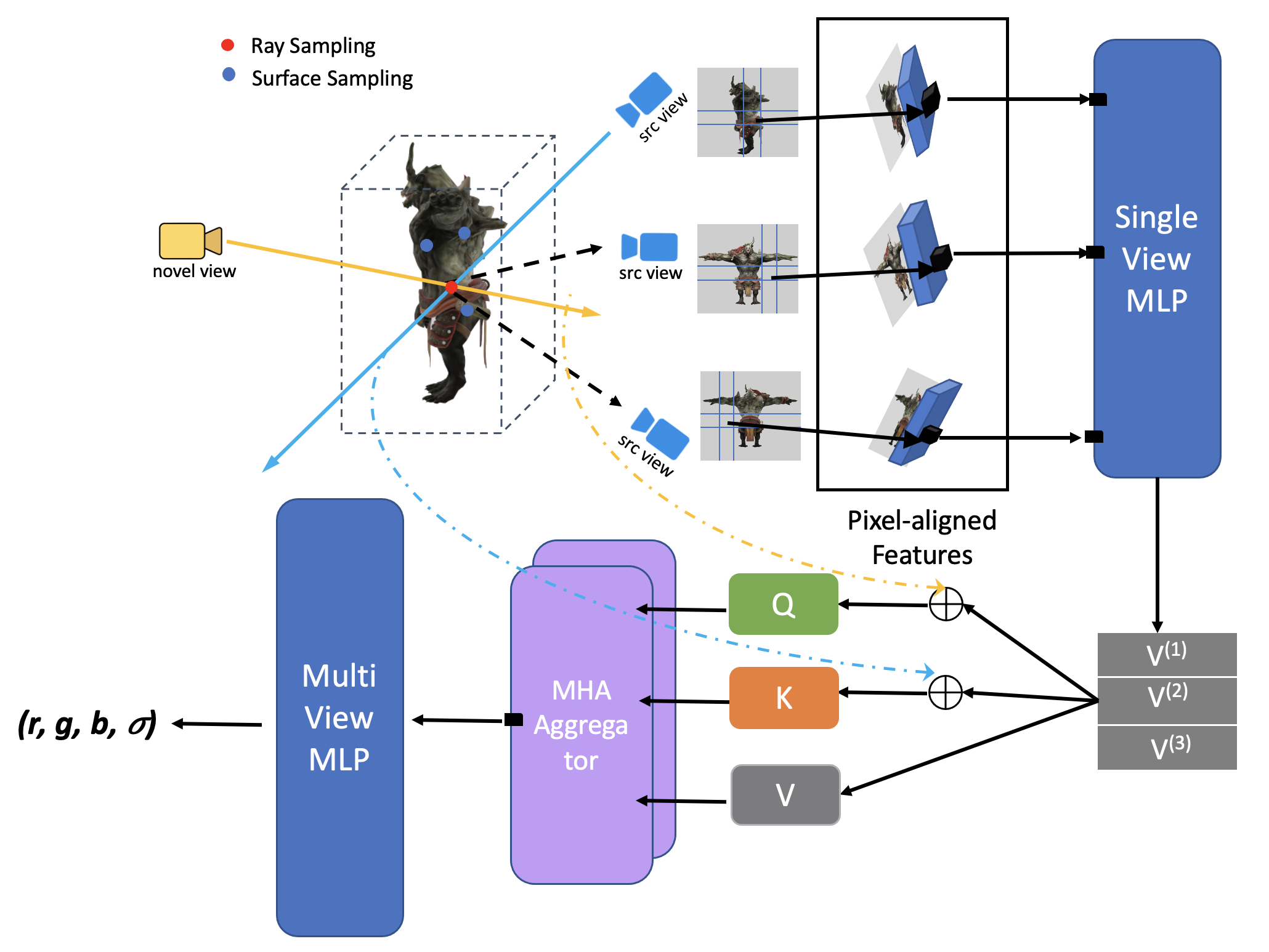}
    \caption{\textnormal{Method Overview: CharNeRF first encodes concept art sketches into pixel aligned feature vectors. During rendering process, feature vectors are extracted and combined through view direction attended self attention layer. Combined features are then fed to the neural radiance field to guide the inference of color and volume density.}}
    \label{fig:architecture}
\end{figure}

\subsection{Overview}
CharNeRF's objective is to reconstruct 3D characters for AR/VR and games using information derived from three consistent turnaround images of the character ( will be referred to as concept art in the rest of the paper). This reconstruction is achieved through the integration of both the visual data and learned knowledge of characters. The system comprises several key components: a convolutional image encoder denoted as $E$, a single-view NeRF MLP labeled as $f_1$, a multi-view NeRF MLP named $f_2$, and a multi-head self-attention-based multi-view feature vector combinator known as $H$.

CharNeRF's pipeline bears resemblance to pixelNeRF~\cite{pixelnerf} but incorporates various modifications to enhance its performance and adapt it to the specific requirements of reconstructing 3D characters.

To effectively utilize the information from concept art images $(I_1, I_2, I_3)$, CharNeRF takes the following steps. First, it independently encodes each sketch into a local feature map, denoted as $W^i = E(I_i)$, where each pixel in the concept art images is transformed into a high-dimensional feature vector capturing shape and color details of nearby pixels.

During the inference of a 3D sample point along with its direction $(x, d)$, CharNeRF projects the 3D point $x$ onto each of the encoded image feature maps $W_i$ using the intrinsic matrix $P_i$. This process extracts feature vectors $W^i(x)$, where $i$ ranges from 1 to 3, corresponding to the three input images. These feature vectors, along with the viewing direction $d$, are then passed through the single-view NeRF MLP $f_1$ to obtain an intermediate feature vector.
\begin{equation}
    V^{(i)} = f_1(W^i(x), d), \ i = 1, ..., 3
\end{equation}
This is done separately and in parallel for each input image. In the end, we obtain $V^{(1)}, V^{(2)}, V^{(3)}$ corresponding to three input sketches. The intermediate feature vectors are passed into the multi-view feature combinator $H$ to obtain a multi-view intermediate feature $V_\text{multi} = H(V^{(1)}, V^{(2)}, V^{(3)})$, which is then used as input to multi-view NeRF MLP $f_2$ that predicts the final colour and density:
\begin{equation}
    f_2(V_\text{multi}, d) = (c, \sigma)
\end{equation}

Both of the NeRF MLPs, namely $f_1$ and $f_2$, adopt a methodology that closely resembles the one outlined in the original NeRF paper~\cite{nerf}. Each of these MLPs is structured with both a coarse and a fine model, and they employ the same ray sampling framework.

In order to enable CharNeRF to learn the characteristics of a specific target category, we train it using a dataset of 3D game characters. Rather than relying solely on the reconstruction of novel views, we take a different approach by directly sampling points from the surfaces of 3D meshes. This strategy serves to guide the neural radiance field in learning the general shape and features specific to the chosen category. Please refer to Figure \ref{fig:architecture} for an overview of our method's architecture.

\subsection{Encode concept art}
The choice of the encoder $E$ holds significant promise for enhancing the quality of information extracted from concept art. In our study, we introduced a novel approach by leveraging the power of double-stacked hourglass encoders, inspired by the success of PiFUHD~\cite{saito2020pifuhd} but with a simplified feature combination. Hourglass networks have demonstrated remarkable effectiveness in extracting poses and keypoints for clothed human figures from 2D images~\cite{pifu}~\cite{saito2020pifuhd}.

Most characters share a fundamental human-like shape but exhibit a wide range of stylistic variations and silhouettes. To capture the intricate shapes of these characters, we deemed it necessary to employ two stacked hourglass networks. Our intuition was to independently encode both high and low-level details from the high and low-resolution versions of the source sketches. While a more complex approach involving an additional MLP for projecting low-dimensional feature vectors before combining them with high-dimensional ones has been suggested~\cite{saito2020pifuhd}, our research led us to a simpler solution – concatenation. This choice not only ensures more stable training but also reduces the risk of encountering gradient vanishing and gradient explosion issues, making the network easier to fine-tune and yielding superior performance.

In the upcoming experiment section, we will delve into the performance of various encoders, shedding more light on our approach's promise and efficacy

\subsection{View direction attended multi-head self attention for feature vector combination}
Average pooling serves as a common technique for merging extracted feature vectors, but it has limitations in terms of learning the importance between these vectors. Our inspiration comes from GeoNeRF \cite{johari2022geonerf}, which introduced a method to aggregate view tokens extracted from cascaded cost volumes of different views using a multi-head self-attention mechanism. In line with this innovative approach, we propose to enhance CharNeRF by fusing the three feature vectors, $V^{(1)}, V^{(2)}, V^{(3)},$ extracted from the concept art sketches through a multi-head self-attention layer. This new approach pays extra attention to the similarity between the query view direction and the source sketch view direction.

In CharNeRF, this is achieved by concatenating the intermediate feature vectors with query view directions, creating a Query Matrix $Q$, and by concatenating the view direction from the estimated camera of each source sketch $I_i$ with the sample point to form a Key Matrix $K$. Mathematically, our self-attention components can be represented as:
\[
\begin{aligned}
    Q &= \text{Stack}(\{ V^{(1)} \oplus {d}, ..., V^{(n)} \oplus {d} \}) 
    \\
    K &= \text{Stack}(\{ V^{(1)} \oplus {d}^{(1)}, ..., V^{(n)} \oplus {d}^{(n)} \}) 
    \\
    V &= \text{Stack}(\{ V^{(1)}, ..., V^{(n)} \}) 
\end{aligned}
\]
, with $n=3$ in our specific case. Before concatenation is applied, intermediate feature vectors and view directions are normalized separately. Additionally, we introduce a learnable scalar to multiply with the view direction, allowing the network to adapt and learn the relative weighting between feature similarity and view direction similarity.

The underlying idea behind this method is to mimic how a real-world 3D modelist might refer to nearby images of concept art to fill in the gaps, enhancing both view consistency across various angles and the quality of rendered results. This innovative approach promises significant improvements in both these aspects.

\subsection{Mix of 2d ray sampling and 3d surface sampling}
\label{surface sampling}
Neural radiance field is typically trained through minimizing the reconstruction difference between rendering and ground truth images from novel angles. This setting allows the network to perform well in novel view synthesis. Nonetheless, pixel colors are estimated through the volumetric rendering which composites colors of all sample points along a ray (see formula \ref{color-composite}), it is natural for neural radiance field to cheat by distributing volume density sparsely along the rays. Many papers propose regularizing loss terms to mitigate this issue ~\cite{kim2022infonerf, wang2022score, melaskyriazi2023realfusion}. Though these approaches tend to improve floating artefact problem, they are unable help improve rendering quality since without extra knowledge provided, the network is blindly penalizing sparse density. We proposed a mix of surface sampling and ray sampling to address this issue. Surface sampling encourages a more centralized density distribution along a ray with higher density near the surface; while ray sampling allows the network to improve novel view rendering quality. We implement surface sampling by first allocating total surface sample points according to the surface area of all sub-meshes due to the fact that characters with complicated shape tend not be air-tight. $n_{\text{mesh}_i} \varpropto  \text{area}_{\text{mesh}_i}$ On each mesh, points are randomly sampled through triangle point picking algorithm ~\cite{weisstein2021triangle}. Points sampled are passed to neural radiance field along with zero view direction vector. The use of zero view direction is based on the assumption of constant lighting where a surface point should have the same color from all view directions. The opaqueness of the surface point $\alpha$ is computed through an estimated $\delta$ value by this simple formula $\delta = \frac{z_\text{far} - z_\text{near} }{n_\text{coarse} + n_\text{fine}}$, where the $z_\text{near}$ and $z_\text{far}$ are the actual depth values. (see formula in Section \ref{integral-estimate} for how $\delta$ is used to estimate integral, note that the normalized depth value t was used in the formula) We observe empirically better rendering results through this method. More analysis will be provided in the evaluation section. (Section \ref{evaluation})

\subsection{Final Loss function}
The final loss contains two terms - reconstruction loss and surface loss. Reconstruction loss is the simple L2 color loss between the ground truth pixel value and the composited color (please refer to Equation \ref{colorcompositing} for color compositing). While the surface loss is L2 color loss of a specific xyz point and L1 loss of the $\alpha$ value. Notice that coarse sampling and fine sampling points are passed to two separate NeRF networks but they will both backpropagate and train on the same image encoder. The formula of the final loss is given below.

\[
    \mathcal{L}_{\text{final}} = \mathcal{L}_{\text{reconstruction}} +  \mathcal{L}_{\text{surface}}
\]

where reconstruction loss is given by:
\[
\begin{aligned}
    \mathcal{L}_\text{reconstruction} 
    &= \mathcal{L}_{\text{reconstruction}, c} + \mathcal{L}_{\text{reconstruction}, f}
    \\
    &= \frac{1}{|\mathcal{R}(\mathbf{P})|} \sum_{\mathbf{r} \in \mathcal{R}(\mathbf{P})}
    \lVert \hat{\mathbf{C}}_c(\mathbf{r}) -  \mathbf{C}(\mathbf{r}) \rVert ^2 _2
    \\ 
    &+ \frac{1}{|\mathcal{R}(\mathbf{P})|} \sum_{\mathbf{r} \in \mathcal{R}(\mathbf{P})}
    \lVert \hat{\mathbf{C}}_f(\mathbf{r}) -  \mathbf{C}(\mathbf{r}) \rVert ^2 _2
\end{aligned}
\]

and surface loss is given by:
\[
\begin{aligned}
    \mathcal{L}_{\text{surface}}
    &= \mathcal{L}_{\text{surface}, c} + \mathcal{L}_{\text{surface}, f}
    \\
    &= \frac{1}{N_p} \Big( \sum_{j=1}^{N_p} \lVert \mathbf{\hat{c}}_{s_j,c} - \mathbf{c}_{s_j} \rVert ^2 _2  + | \hat\alpha_{s_j,c} - \alpha_{s_j} | \Big)
    \\
    &+ \frac{1}{N_p} \Big( \sum_{j=1}^{N_p} \lVert \mathbf{\hat{c}}_{s_j,f} - \mathbf{c}_{s_j} \rVert ^2 _2  + | \hat\alpha_{s_j,f} - \alpha_{s_j} | \Big)
\end{aligned}
\]

\subsection{Mesh Reconstruction}
\begin{figure}[hbt!]
    \centering
    \includegraphics[width=9cm]{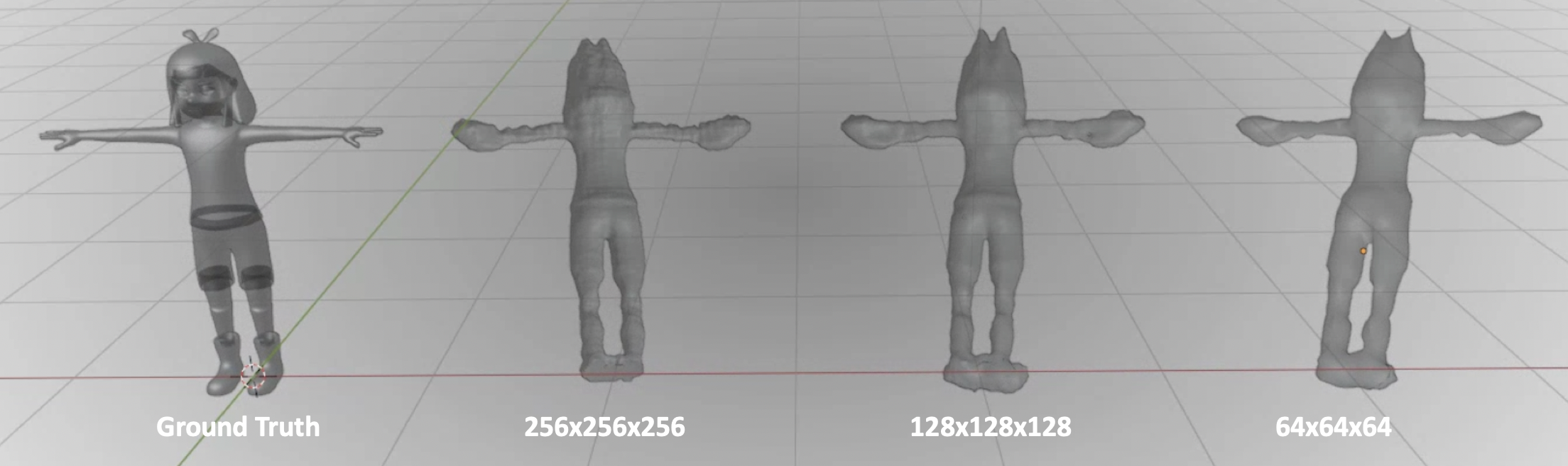}
    \caption{The marching cube algorithm is applied with three distinct resolutions to the same object, resulting in three different mesh representations that can be regarded as three different levels of detail. The polygon and vertex counts for each of these representations are provided above}
    \label{fig:method:lodMesh}
\end{figure}

In the context of mesh reconstruction from NeRF, the customary procedure entails the application of the marching cubes algorithm \cite{marchingCube} to the estimated volume density cube, with the size of this cube governing the resolution of the resulting mesh. This approach offers a significant advantage, particularly when varying levels of detail (LOD) are required for the same characters. Different mesh resolutions can be generated to suit diverse requirements. See Figure \ref{fig:method:lodMesh}, and Table \ref{table:lod}.

\begin{table}[h!]
\centering
\caption{\label{table:lod}The number of vertices and polygons for meshes generated using different resolutions of voxel grids.}
\begin{tabular}{|l|l|l|l|}
\hline
Resolution   & 64x64x64   & 128x128x128   & 256x256x256   \\ \hline
No. Vertices & 1822 & 7646  & 49582 \\ \hline
No. Polygons & 3640 & 15288 & 98972 \\ \hline
\end{tabular}
\end{table}

\begin{figure}[h]
    \centering
    \includegraphics[width=\columnwidth]{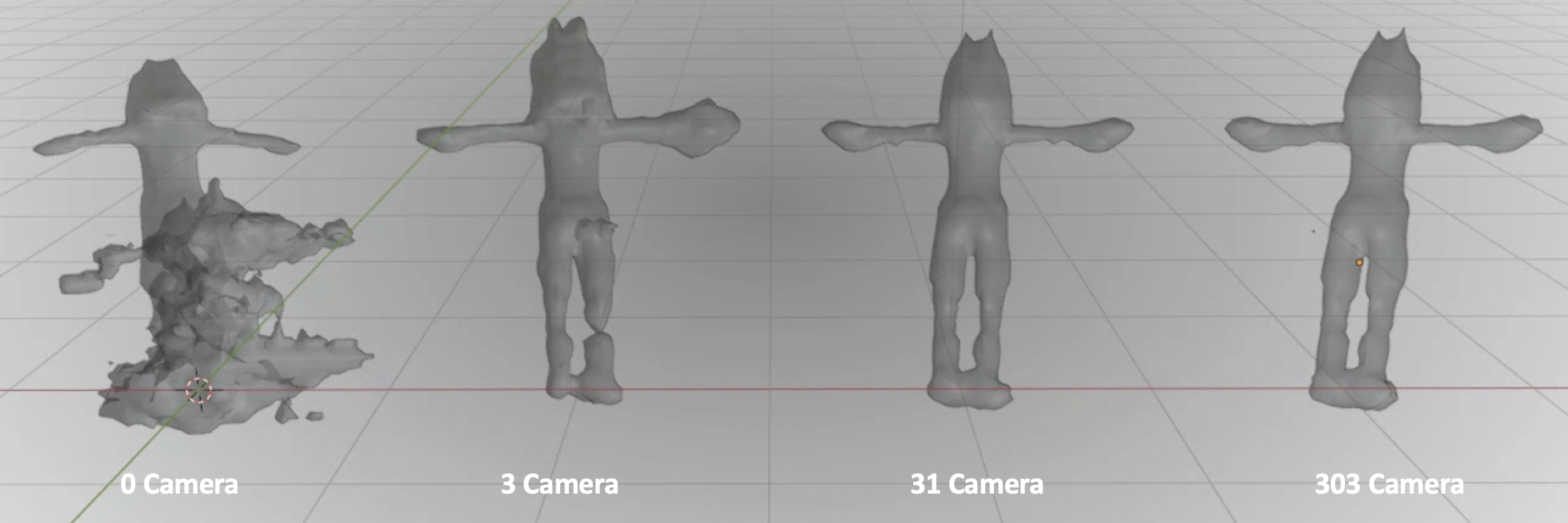}
    \caption{The estimated densities have been computed as an average across varying numbers of camera angles. It is observed that an overall positive correlation exists between mesh quality and the number of camera angles employed. Mesh constructed using resolution 64x64x64}
    \label{fig:methods:numOfCamMesh}
\end{figure}

However, it's essential to note that in the original NeRF paper \cite{nerf}, the process of inferring volume densities for mesh reconstruction involves fixing the viewing directions to zero. Regrettably, this method effectively nullifies the learned feature combination mechanism, leading to empirically unsatisfactory results. In light of these limitations, we advocate a novel approach that entails providing multiple camera angles and subsequently averaging the estimated volume densities. Our empirical findings demonstrate that the employment of a greater number of camera angles enhances the quality of the resulting mesh. See Figure \ref{fig:methods:numOfCamMesh}.

\section{Experiments}
\subsection{Implementation Details}
\label{implementation details}
\textbf{Network Structure}
In our experimental investigations, we explored various configurations of single-view NeRF MLP blocks preceding the feature combination step, ranging from immediate combination after feature extraction to integration after all MLP blocks. Ultimately, our results revealed that employing three single-view MLP blocks provided an optimal configuration. Interestingly, our high-level network architecture bears similarities to PixelNeRF~\cite{pixelnerf}; however, our model exhibits superior performance in the context of 3D virtual character modeling (for quantitative evaluation, refer to Section \ref{evaluation}).

\begin{figure}[h]
    \centering
    \includegraphics[width=9cm]{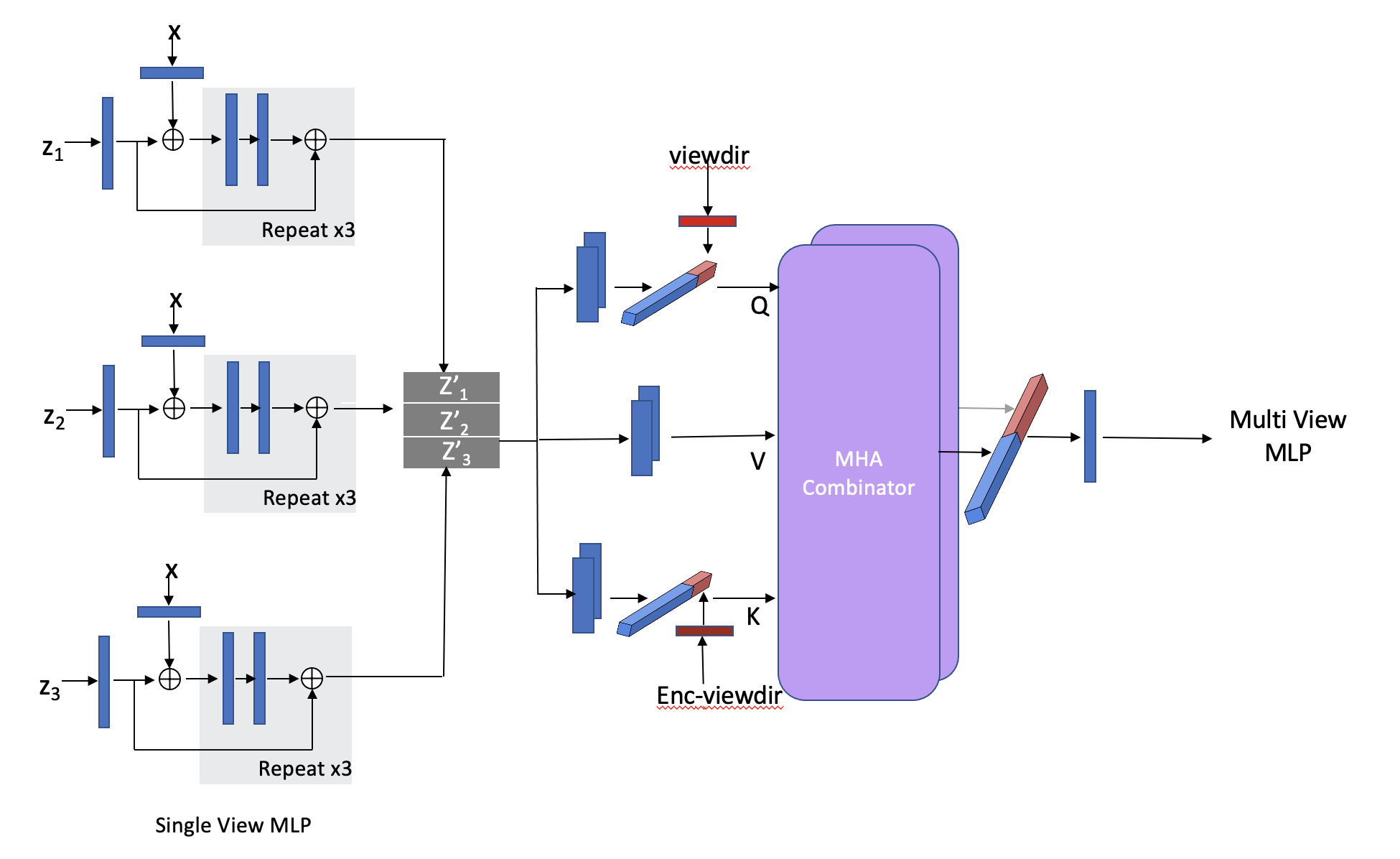}
    \caption{The overall network structure. \textnormal{$z_i$ represents the feature vectors extracted from the feature maps, and $x$ is positional encoded 3D coordinates and view direction. Both $x$ and $z_i$ are projected through an independent linear layer before their combination. Three combined information are passed through three single-view MLP blocks in parallel before being combined through a view direction attended multi head self attention combinator.}}
    \label{fig:mha_network}
\end{figure}

While experimenting with diverse per-scene optimization strategies, we observed that exclusively updating the late MLP blocks allowed the network to adapt towards specific instances. Although we did not identify an efficient per-scene optimization method within the scope of our research, this discovery presents an intriguing avenue for future exploration. It suggests a potential synthesis of regressive and generative approaches, where the generative method could be employed for character-specific optimization by training over the late MLP blocks. See Figure \ref{fig:mha_network} for the network structure of CharNeRF.

\textbf{Dataset Collection}
We began by collecting 75 3D characters in fbx format from TurboSquid, as found at \footnote{\href{https://www.turbosquid.com/3d-model/free/character/fbx?page_num=2}{https://www.turbosquid.com/3d-model/free/character/fbx?page\_num=2}}. For each character, the initial step involves resizing the character's height to 2 meters, after which we generate 300 images by evenly sampling camera positions across the upper hemisphere that encompasses the character. The even distribution of camera positions is achieved using the Fibonacci Lattice method. Each camera's 4x4 "camera to world" matrix is saved as a txt file. Additionally, we create a unified JSON file containing intrinsic information, encompassing focal length, center pixel location, distortion, width, height, and AABB scale. To facilitate surface sampling, we also extract texture images, material files, and object files for each character. Furthermore, we produce 2D depth maps for each image, which are later employed in experiments involving ground truth depth as a prior. This entire process is implemented procedurally through a Blender Python Script.\footnote{\href{https://docs.blender.org/manual/en/latest/advanced/scripting/introduction.html}{https://docs.blender.org/manual/en/latest/advanced/scripting/introduction.html}}

We further collect characters from Mixamo \footnote{\href{https://www.mixamo.com/}{https://www.mixamo.com/}}, and copyright free concept art sketches from Pinterest \footnote{\href{https://www.pinterest.com/}{https://www.pinterest.com/}}, Artstation \footnote{\href{https://www.artstation.com/?sort_by=community&dimension=all}{https://www.artstation.com/?sort\_by=community\&dimension=all}} and DevianArt \footnote{\href{https://www.deviantart.com/}{https://www.deviantart.com/}} for validation and evaluation. Overall, we have 75, 11, and 12 characters for training, validation, and testing respectively.

\textbf{Hyperparameter selection}
The selection of $\lambda$ for surface loss and reconstruction loss affects the trained model. We use $\lambda_\text{reconstruction} = 1.0$ and $\lambda_\text{surface} = 0.1$ for our final model. In the early phase of training, we find small number of sample points per iteration helps neural radiance field learn the general shape fast. We chooses $n_\text{reconstruction, ray}=1000$ and $n_\text{surface} = 3000$. Along each ray, coarse sampling samples 64 points, while fine sampling samples extra 128 points. While after 4 days of training, we increase the $n_\text{reconstruction, ray}$ to 10000 and $n_\text{surface}$ to 10000. The training process goes on for another 3 days.

\subsection{Evaluation}
\label{evaluation}

\textbf{Evaluation Metrics} 
We use three evaluation metrics for quantitative analysis of model performances: Peak Signal-to-noise Ratio (PSNR) and Structural Similarity Index (SSIM)~\cite{wang_image_2004} metrics for assessing the quality of images rendered by our model against the ground truth, and Learned Perceptual Image Patch Similarity (LPIPS)~\cite{zhang_unreasonable_2018} as a better score that reflects a human's perception of similarities between the rendered and ground truth.

\textbf{Baselines}
We choose pixelNeRF~\cite{pixelnerf} as baseline for our model. Furthermore, we include several baselines by removing or modifying components in our final model proposed for ablation analysis. Specifically, we include: 1) our model using one Stacked Hourglass network encoder (instead of using two parallel coarse and fine encoders), 2) our model without MHA aggregator $H$, 3) our model without MHA Aggregator and with extra depth ground-truth for loss term, and 4) our model with a simple Self-Attention aggregator without view information. In total, we have six experiments: our final model CharNeRF and five baselines.

\begin{figure}[h]
    \centering
    \includegraphics[width=8cm]{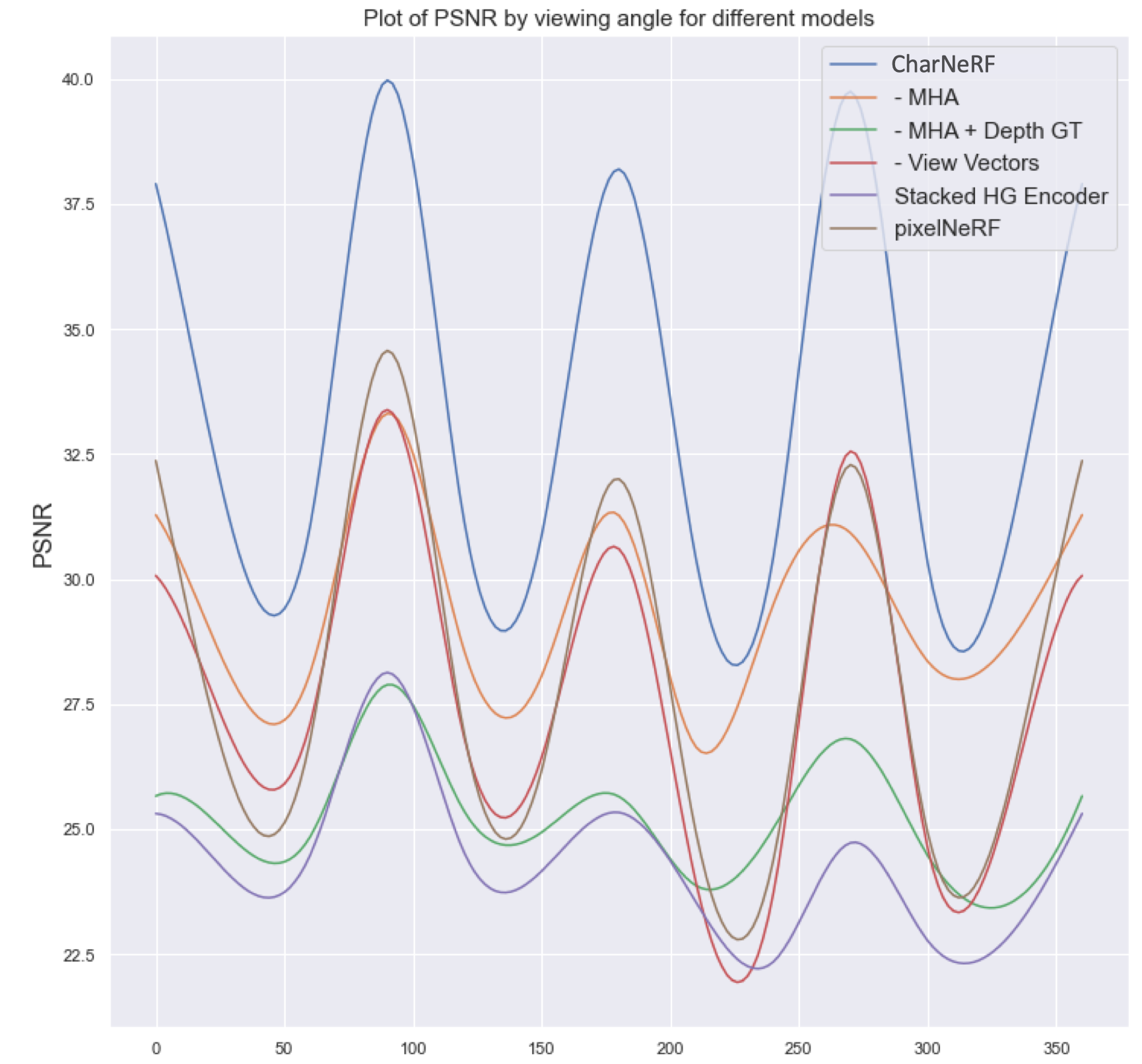}
    \caption{Plot of average PSNR by viewing angles for different models \textnormal{where the higher the score the better and CharNeRF (in blue) gives the best results regardless of viewing angles.}}
    \label{fig:eval:ablation-boxplot-psnr}
\end{figure}

\textbf{Quantitative Results}
We present the average metrics results across all viewing angles of all test dataset characters in Table~\ref{tab:eval:metrics}. CharNeRF gives the best results in terms of all three evaluation metrics. We are also aware that the quality of rendered image varies greatly across the rotation angles - if the target view is novel and has a great angle difference with any of source views, the result will be less ideal. As such we plot the PSNR values for different viewing angles summarising the distributions of PSNR values across different views by experiment, which are shown in Figure~\ref{fig:eval:ablation-boxplot-psnr}. 


\begin{table}[]
\begin{center}
\caption{Evaluation metrics on the test dataset for different models\textnormal{ Note that CharNeRF outperforms the others for PSNR (higher the better), SSIM (higher the better), and LPIPS (lower the better)}} 
\begin{tabular}{lllll}
\hline
    &           & PSNR $\uparrow$   & SSIM $\uparrow$   & LPIPS $\downarrow$ \\
    \hline
    & pixelNeRF & 29.2              & 0.973             & 0.071 \\
    \hline
\multirow{5}{*}{\rotatebox{90}{Our Models}} 
    & 1) Using Stacked Hourglass
                & 25.1              & 0.955             & 0.147 \\
    & 2) No MHA & 29.8              & 0.976             & 0.073 \\
    & 3) No MHA, with depth GT 
                & 25.6              & 0.966             & 0.128 \\
    & 4) No view directions in MHA
                & 28.7              & 0.974             & 0.072 \\
    & 5) CharNeRF
                & \textbf{34.3}     & \textbf{0.988}    & \textbf{0.031}
    \\
\hline
\end{tabular}

\label{tab:eval:metrics}
\end{center}
\end{table}

\begin{figure}[h]
    \centering
    \includegraphics[width=8cm]{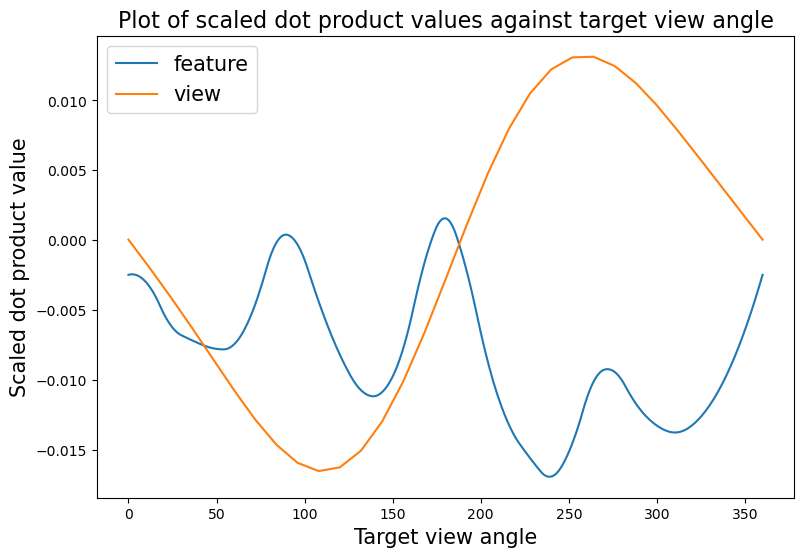}
    \caption{Plot of mean scaled dot product values against query view direction \textnormal{on a test image using CharNeRF. Note that the final scaled dot product value is a sum of the two components, i.e. feature vector (labelled "feature") and view direction (labelled "view")}}
    \label{fig:eval:dotprod}
\end{figure}

\textbf{Analysis of Multi-view Feature Vector Combination}
We have also conducted experiments to analyse whether our proposed method of using both intermediate feature vectors and view direction information for multi-head self attention is meaningful. For each pixel per head, we first extract the scaled dot product value, which is a sum of the dot product of the feature sub-vectors and that of the view direction vectors. Then, we take the average across all pixels and heads in one image to get a mean scaled dot product value contributed to feature vectors and one contributed to view direction information. We repeat the process for all $360^\circ$ views of the same character and plot the scaled dot product components in Figure~\ref{fig:eval:dotprod}. 

The observed results closely align with our theoretical framework. The two lines exhibit variations based on query view directions, each contributing differently to the final scaled dot product value in self-attention for multi-view combination due to their distinct learnable weights. Specifically, the line corresponding to feature vectors demonstrates significant and frequent fluctuations across query view directions. It consistently attains high values when the angle aligns closely with any of the source views, such as $0^\circ$ ($360^\circ$), $90^\circ$, and $180^\circ$, while registering minimal values at angles between any two input images' views, like $45^\circ$. Intuitively, this behavior indicates the network's inclination to use feature similarity as a guide when combining features for novel views close to the source views.

Moreover, given that only one side image is provided, even though the model can infer the other side due to symmetry in most sketches, the plot exhibits lower values from $180^\circ$ to $360^\circ$ compared to the range from $0^\circ$ to $180^\circ$. Notably, the scaled dot product values related to view direction information trace a smooth sinusoidal curve, resembling a multiplication of the simple function $y = f(x) = \cos x + \cos(x - 90^\circ) + \cos(x - 180^\circ)$ for $x \in [0^\circ, 360^\circ]$. This theoretical curve represents the sum of dot products between the query and each input source view, providing insight into how the network utilizes view direction information in the context of our proposed methodology.

\begin{figure}[h]
    \centering
    \includegraphics[width=\columnwidth]{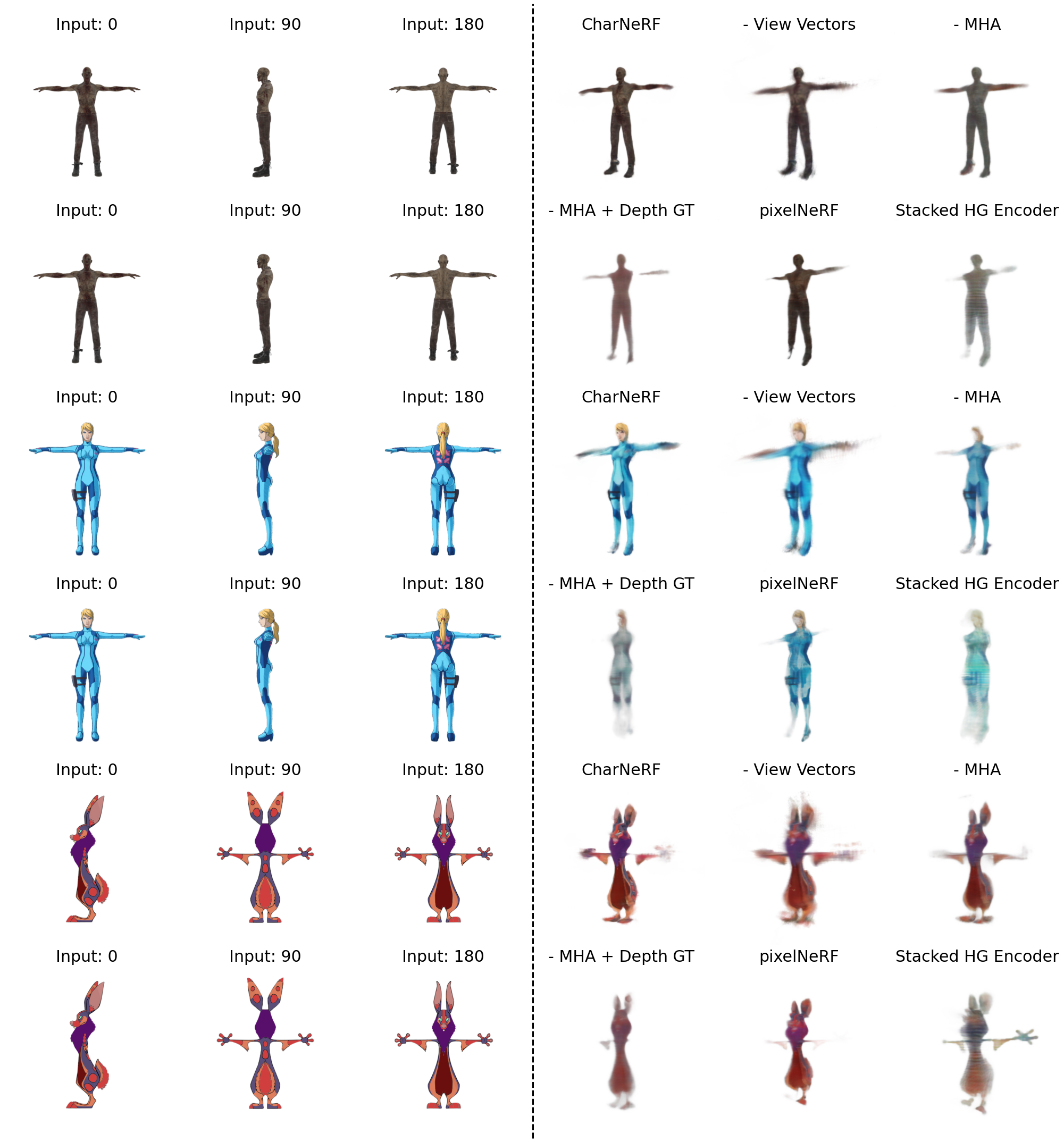}
    \caption{Visualisation for the same character and same views by different models \textnormal{where every two row refers to one character and the left three columns are the input images.}}
    \label{fig:eval:ablation-img-matrix}
\end{figure}

\textbf{Qualitative Results}
For qualitative results, we randomly choose a few characters from the Mixamo dataset as well as several copy-free concept art sketches from Artstation and we render each character from 3 image inputs for each of our baseline and the final model. The results are shown in Figure~\ref{fig:eval:ablation-img-matrix}. Visually, our final model CharNeRF gives the best result as it is able to reconstruct the colour and shape most accurately and the rendered image is least blurry in terms of peripheral body parts of the characters, which are challenging regions of the character body for novel view synthesis.

\section{Future Extension}
Recent innovations have propelled pre-trained 2D diffusion models into the realm of 3D, unlocking a rich avenue for exploration \cite{poole2022dreamfusion, wang2022score}. Subsequent studies have further expanded this frontier, venturing into image-based reconstruction. Some researchers have achieved this by employing textual inversion to train instance-specific embeddings for 2D pre-trained diffusion models \cite{melaskyriazi2023realfusion}, while others have opted for training image-conditioned diffusion models \cite{liu2023zero1to3, watson2022novel}. This trajectory capitalizes on the potent generative capabilities of diffusion models as priors for neural radiance fields, effectively addressing the challenge of requiring extensive training data for regressive approaches. As mentioned in Section \ref{implementation details}, our proposed CharNeRF framework can be easily fine-tuned for specific instances by enhancing the second MLP. Interestingly, we discovered that NerfDiff \cite{gu2023nerfdiff} shares a similar concept, distilling the knowledge of diffusion models to refine neural radiance fields for specific instances. This approach seamlessly integrates with our model, offering a ready-made extension and paving the way for intriguing future explorations. Moreover, given that concept art often provides detailed views of occluded body parts in real-world scenarios, extracting insights from such information to enhance predicted meshes holds substantial value for the industry as it provides editability for the mesh reconstruction from concept art. In summary, delving into generative approaches using diffusion models and mastering the utilization of additional information represent captivating directions for extending our proposed topic and methodology.

\section{Conclusion}
To conclude, the paper attempts to address one challenging problem in 3D computer vision that has great values in AR/VR/Game applications, which is to use NeRF to construct a 3D representation for a 3D character from concept art. Our final model proposed, CharNeRF, is able to generate good results from such sparse image inputs thanks to view direction attended multi-head self attention component used for combining information from different input views. However, there are still limitations to CharNeRF such as inability to generate fine details of the character especially at the hand region from novel angles that are far from source images, which is largely due to the high variation of the dataset and the limited number of data that we can retrieve, and this can be a potential area of work for future researchers.

\bigbreak
\noindent
\textbf{Acknowledgements}
This work is supported by the Singapore Ministry of Education Academic Research grant T1 251RES2205, “Realtime Distributed Hybrid Rendering with 5G Edge Computing for
Realistic Graphics in Mobile Games and Metaverse Applications”.

\bibliography{conference_101719}

\begin{thebibliography}{10}
\providecommand{\url}[1]{#1}
\csname url@samestyle\endcsname
\providecommand{\newblock}{\relax}
\providecommand{\bibinfo}[2]{#2}
\providecommand{\BIBentrySTDinterwordspacing}{\spaceskip=0pt\relax}
\providecommand{\BIBentryALTinterwordstretchfactor}{4}
\providecommand{\BIBentryALTinterwordspacing}{\spaceskip=\fontdimen2\font plus
\BIBentryALTinterwordstretchfactor\fontdimen3\font minus \fontdimen4\font\relax}
\providecommand{\BIBforeignlanguage}[2]{{%
\expandafter\ifx\csname l@#1\endcsname\relax
\typeout{** WARNING: IEEEtran.bst: No hyphenation pattern has been}%
\typeout{** loaded for the language `#1'. Using the pattern for}%
\typeout{** the default language instead.}%
\else
\language=\csname l@#1\endcsname
\fi
#2}}
\providecommand{\BIBdecl}{\relax}
\BIBdecl

\bibitem{summerville_procedural_2018}
\BIBentryALTinterwordspacing
A.~Summerville, S.~Snodgrass, M.~Guzdial, C.~Holmgård, A.~K. Hoover, A.~Isaksen, A.~Nealen, and J.~Togelius, ``Procedural {Content} {Generation} via {Machine} {Learning} ({PCGML}),'' May 2018, arXiv:1702.00539 [cs]. [Online]. Available: \url{http://arxiv.org/abs/1702.00539}
\BIBentrySTDinterwordspacing

\bibitem{highQualitySteamable}
\BIBentryALTinterwordspacing
A.~Collet, M.~Chuang, P.~Sweeney, D.~Gillett, D.~Evseev, D.~Calabrese, H.~Hoppe, A.~Kirk, and S.~Sullivan, ``High-quality streamable free-viewpoint video,'' \emph{ACM Trans. Graph.}, vol.~34, no.~4, jul 2015. [Online]. Available: \url{https://doi.org/10.1145/2766945}
\BIBentrySTDinterwordspacing

\bibitem{theRelightables}
\BIBentryALTinterwordspacing
K.~Guo, P.~Lincoln, P.~Davidson, J.~Busch, X.~Yu, M.~Whalen, G.~Harvey, S.~Orts-Escolano, R.~Pandey, J.~Dourgarian, D.~Tang, A.~Tkach, A.~Kowdle, E.~Cooper, M.~Dou, S.~Fanello, G.~Fyffe, C.~Rhemann, J.~Taylor, P.~Debevec, and S.~Izadi, ``The relightables: Volumetric performance capture of humans with realistic relighting,'' \emph{ACM Trans. Graph.}, vol.~38, no.~6, nov 2019. [Online]. Available: \url{https://doi.org/10.1145/3355089.3356571}
\BIBentrySTDinterwordspacing

\bibitem{saito2020pifuhd}
S.~Saito, T.~Simon, J.~Saragih, and H.~Joo, ``Pifuhd: Multi-level pixel-aligned implicit function for high-resolution 3d human digitization,'' in \emph{CVPR}, 2020.

\bibitem{icon}
Y.~Xiu, J.~Yang, D.~Tzionas, and M.~J. Black, ``{ICON}: {I}mplicit {C}lothed humans {O}btained from {N}ormals,'' in \emph{Proceedings of the IEEE/CVF Conference on Computer Vision and Pattern Recognition (CVPR)}, June 2022, pp. 13\,296--13\,306.

\bibitem{AnimationReadyClothedHuman}
\BIBentryALTinterwordspacing
T.~He, Y.~Xu, S.~Saito, S.~Soatto, and T.~Tung, ``{ARCH++:} animation-ready clothed human reconstruction revisited,'' \emph{CoRR}, vol. abs/2108.07845, 2021. [Online]. Available: \url{https://arxiv.org/abs/2108.07845}
\BIBentrySTDinterwordspacing

\bibitem{siclope}
\BIBentryALTinterwordspacing
R.~Natsume, S.~Saito, Z.~Huang, W.~Chen, C.~Ma, H.~Li, and S.~Morishima, ``Siclope: Silhouette-based clothed people,'' \emph{CoRR}, vol. abs/1901.00049, 2019. [Online]. Available: \url{http://arxiv.org/abs/1901.00049}
\BIBentrySTDinterwordspacing

\bibitem{smpl}
\BIBentryALTinterwordspacing
M.~Loper, N.~Mahmood, J.~Romero, G.~Pons-Moll, and M.~J. Black, ``Smpl: A skinned multi-person linear model,'' \emph{ACM Trans. Graph.}, vol.~34, no.~6, nov 2015. [Online]. Available: \url{https://doi.org/10.1145/2816795.2818013}
\BIBentrySTDinterwordspacing

\bibitem{ghum}
H.~Xu, E.~G. Bazavan, A.~Zanfir, W.~T. Freeman, R.~Sukthankar, and C.~Sminchisescu, ``Ghum \& ghuml: Generative 3d human shape and articulated pose models,'' in \emph{Proceedings of the IEEE/CVF Conference on Computer Vision and Pattern Recognition}, 2020, pp. 6184--6193.

\bibitem{scape}
\BIBentryALTinterwordspacing
D.~Anguelov, P.~Srinivasan, D.~Koller, S.~Thrun, J.~Rodgers, and J.~Davis, ``Scape: Shape completion and animation of people,'' \emph{ACM Trans. Graph.}, vol.~24, no.~3, p. 408–416, jul 2005. [Online]. Available: \url{https://doi.org/10.1145/1073204.1073207}
\BIBentrySTDinterwordspacing

\bibitem{nerf}
\BIBentryALTinterwordspacing
B.~Mildenhall, P.~P. Srinivasan, M.~Tancik, J.~T. Barron, R.~Ramamoorthi, and R.~Ng, ``Nerf: Representing scenes as neural radiance fields for view synthesis,'' \emph{CoRR}, vol. abs/2003.08934, 2020. [Online]. Available: \url{https://arxiv.org/abs/2003.08934}
\BIBentrySTDinterwordspacing

\bibitem{Deephuman}
\BIBentryALTinterwordspacing
Z.~Zheng, T.~Yu, Y.~Wei, Q.~Dai, and Y.~Liu, ``Deephuman: 3d human reconstruction from a single image,'' 2019. [Online]. Available: \url{https://arxiv.org/abs/1903.06473}
\BIBentrySTDinterwordspacing

\bibitem{bodyNet}
\BIBentryALTinterwordspacing
G.~Varol, D.~Ceylan, B.~C. Russell, J.~Yang, E.~Yumer, I.~Laptev, and C.~Schmid, ``Bodynet: Volumetric inference of 3d human body shapes,'' \emph{CoRR}, vol. abs/1804.04875, 2018. [Online]. Available: \url{http://arxiv.org/abs/1804.04875}
\BIBentrySTDinterwordspacing

\bibitem{blockNerf}
\BIBentryALTinterwordspacing
M.~Tancik, V.~Casser, X.~Yan, S.~Pradhan, B.~Mildenhall, P.~P. Srinivasan, J.~T. Barron, and H.~Kretzschmar, ``Block-nerf: Scalable large scene neural view synthesis,'' 2022. [Online]. Available: \url{https://arxiv.org/abs/2202.05263}
\BIBentrySTDinterwordspacing

\bibitem{instant-ngp}
\BIBentryALTinterwordspacing
T.~Müller, A.~Evans, C.~Schied, and A.~Keller, ``Instant neural graphics primitives with a multiresolution hash encoding,'' \emph{{ACM} Transactions on Graphics}, vol.~41, no.~4, pp. 1--15, jul 2022. [Online]. Available: \url{https://doi.org/10.1145%2F3528223.3530127}
\BIBentrySTDinterwordspacing

\bibitem{fridovich-keil_plenoxels_2022}
S.~Fridovich-Keil, A.~Yu, M.~Tancik, Q.~Chen, B.~Recht, and A.~Kanazawa, ``Plenoxels: {Radiance} {Fields} {Without} {Neural} {Networks},'' in \emph{Proceedings of the {IEEE}/{CVF} {Conference} on {Computer} {Vision} and {Pattern} {Recognition}}, 2022, pp. 5501--5510.

\bibitem{keypointNerf}
M.~Mihajlovic, A.~Bansal, M.~Zollhoefer, S.~Tang, and S.~Saito, ``{KeypointNeRF}: Generalizing image-based volumetric avatars using relative spatial encoding of keypoints,'' in \emph{European conference on computer vision}, 2022.

\bibitem{portraitNerf}
\BIBentryALTinterwordspacing
C.~Gao, Y.~Shih, W.~Lai, C.~Liang, and J.~Huang, ``Portrait neural radiance fields from a single image,'' \emph{CoRR}, vol. abs/2012.05903, 2020. [Online]. Available: \url{https://arxiv.org/abs/2012.05903}
\BIBentrySTDinterwordspacing

\bibitem{pixelnerf}
A.~Yu, V.~Ye, M.~Tancik, and A.~Kanazawa, ``pixelnerf: Neural radiance fields from one or few images,'' 2021.

\bibitem{johari2022geonerf}
M.~M. Johari, Y.~Lepoittevin, and F.~Fleuret, ``Geonerf: Generalizing nerf with geometry priors,'' 2022.

\bibitem{humannerf}
C.-Y. Weng, B.~Curless, P.~P. Srinivasan, J.~T. Barron, and I.~Kemelmacher-Shlizerman, ``Humannerf: Free-viewpoint rendering of moving people from monocular video,'' 2022.

\bibitem{su_-nerf_2021}
S.-Y. Su, F.~Yu, M.~Zollhöfer, and H.~Rhodin, ``A-nerf: {Articulated} neural radiance fields for learning human shape, appearance, and pose,'' \emph{Advances in Neural Information Processing Systems}, vol.~34, pp. 12\,278--12\,291, 2021.

\bibitem{kim2022infonerf}
M.~Kim, S.~Seo, and B.~Han, ``Infonerf: Ray entropy minimization for few-shot neural volume rendering,'' 2022.

\bibitem{niemeyer2022regnerf}
M.~Niemeyer, J.~T. Barron, B.~Mildenhall, M.~S. Sajjadi, A.~Geiger, and N.~Radwan, ``Regnerf: Regularizing neural radiance fields for view synthesis from sparse inputs,'' in \emph{Proceedings of the IEEE/CVF Conference on Computer Vision and Pattern Recognition}, 2022, pp. 5480--5490.

\bibitem{self-attention}
A.~Vaswani, N.~Shazeer, N.~Parmar, J.~Uszkoreit, L.~Jones, A.~N. Gomez, L.~Kaiser, and I.~Polosukhin, ``Attention is all you need,'' 2017.

\bibitem{endToEndRecovery}
\BIBentryALTinterwordspacing
A.~Kanazawa, M.~J. Black, D.~W. Jacobs, and J.~Malik, ``End-to-end recovery of human shape and pose,'' \emph{CoRR}, vol. abs/1712.06584, 2017. [Online]. Available: \url{http://arxiv.org/abs/1712.06584}
\BIBentrySTDinterwordspacing

\bibitem{zheng_pamir_2022}
Z.~Zheng, T.~Yu, Y.~Liu, and Q.~Dai, ``{PaMIR}: {Parametric} {Model}-{Conditioned} {Implicit} {Representation} for {Image}-{Based} {Human} {Reconstruction},'' \emph{IEEE Transactions on Pattern Analysis and Machine Intelligence}, vol.~44, no.~6, pp. 3170--3184, Jun. 2022, conference Name: IEEE Transactions on Pattern Analysis and Machine Intelligence.

\bibitem{pifu}
\BIBentryALTinterwordspacing
S.~Saito, Z.~Huang, R.~Natsume, S.~Morishima, A.~Kanazawa, and H.~Li, ``Pifu: Pixel-aligned implicit function for high-resolution clothed human digitization,'' \emph{CoRR}, vol. abs/1905.05172, 2019. [Online]. Available: \url{http://arxiv.org/abs/1905.05172}
\BIBentrySTDinterwordspacing

\bibitem{liu2016deepfashion}
Z.~Liu, P.~Luo, S.~Qiu, X.~Wang, and X.~Tang, ``Deepfashion: Powering robust clothes recognition and retrieval with rich annotations,'' in \emph{Proceedings of the IEEE conference on computer vision and pattern recognition}, 2016, pp. 1096--1104.

\bibitem{hornik_multilayer_1989}
K.~Hornik, M.~Stinchcombe, and H.~White, ``Multilayer feedforward networks are universal approximators,'' \emph{Neural networks}, vol.~2, no.~5, pp. 359--366, 1989, publisher: Elsevier.

\bibitem{wang2022score}
H.~Wang, X.~Du, J.~Li, R.~A. Yeh, and G.~Shakhnarovich, ``Score jacobian chaining: Lifting pretrained 2d diffusion models for 3d generation,'' 2022.

\bibitem{melaskyriazi2023realfusion}
L.~Melas-Kyriazi, C.~Rupprecht, I.~Laina, and A.~Vedaldi, ``Realfusion: 360{$\deg$} reconstruction of any object from a single image,'' 2023.

\bibitem{weisstein2021triangle}
\BIBentryALTinterwordspacing
E.~W. Weisstein, ``Triangle point picking,'' MathWorld--A Wolfram Web Resource, 2021. [Online]. Available: \url{https://mathworld.wolfram.com/TrianglePointPicking.html}
\BIBentrySTDinterwordspacing

\bibitem{marchingCube}
\BIBentryALTinterwordspacing
W.~E. Lorensen and H.~E. Cline, ``Marching cubes: A high resolution 3d surface construction algorithm,'' \emph{SIGGRAPH Comput. Graph.}, vol.~21, no.~4, p. 163–169, aug 1987. [Online]. Available: \url{https://doi.org/10.1145/37402.37422}
\BIBentrySTDinterwordspacing

\bibitem{wang_image_2004}
Z.~Wang, A.~C. Bovik, H.~R. Sheikh, and E.~P. Simoncelli, ``Image quality assessment: from error visibility to structural similarity,'' \emph{IEEE transactions on image processing}, vol.~13, no.~4, pp. 600--612, 2004, publisher: IEEE.

\bibitem{zhang_unreasonable_2018}
R.~Zhang, P.~Isola, A.~A. Efros, E.~Shechtman, and O.~Wang, ``The unreasonable effectiveness of deep features as a perceptual metric,'' in \emph{Proceedings of the {IEEE} conference on computer vision and pattern recognition}, 2018, pp. 586--595.

\bibitem{poole2022dreamfusion}
B.~Poole, A.~Jain, J.~T. Barron, and B.~Mildenhall, ``Dreamfusion: Text-to-3d using 2d diffusion,'' 2022.

\bibitem{liu2023zero1to3}
R.~Liu, R.~Wu, B.~V. Hoorick, P.~Tokmakov, S.~Zakharov, and C.~Vondrick, ``Zero-1-to-3: Zero-shot one image to 3d object,'' 2023.

\bibitem{watson2022novel}
D.~Watson, W.~Chan, R.~Martin-Brualla, J.~Ho, A.~Tagliasacchi, and M.~Norouzi, ``Novel view synthesis with diffusion models,'' 2022.

\bibitem{gu2023nerfdiff}
J.~Gu, A.~Trevithick, K.-E. Lin, J.~Susskind, C.~Theobalt, L.~Liu, and R.~Ramamoorthi, ``Nerfdiff: Single-image view synthesis with nerf-guided distillation from 3d-aware diffusion,'' 2023.

\end{thebibliography}
\bibliographystyle{IEEEtran}

\vspace{12pt}

\end{document}